\documentclass[10pt,journal,compsoc]{IEEEtran}
\usepackage{ragged2e}

\usepackage{verbatim}
\usepackage{ulem}
\usepackage{times}
\usepackage{epsfig}
\usepackage{graphicx}
\usepackage{amsmath}
\usepackage{amssymb}
\usepackage{mathrsfs}
\usepackage{multirow}
\usepackage{makecell}
\usepackage{epstopdf}
\usepackage{amsthm}
\usepackage{bbm}
\usepackage{algorithm}
\usepackage{color} 
\usepackage{enumitem}
\usepackage{booktabs}
\usepackage{algorithmicx}
\usepackage{algpseudocode}
\usepackage{cleveref}

\ifCLASSOPTIONcompsoc
  \usepackage[nocompress]{cite}
\else
  \usepackage{cite}
\fi

\usepackage[utf8]{inputenc}
\usepackage[T1]{fontenc}
\usepackage{listings}
\usepackage{xcolor} 
\ifCLASSINFOpdf
\else
\fi
\title{
Causal Disentanglement for Robust Long-tail Medical Image Generation}

\author{
        Weizhi Nie,~
        Zichun Zhang,~
        Weijie Wang$^*$,~
        Bruno Lepri,~
        Anan Liu,~
        Nicu Sebe

\IEEEcompsocitemizethanks{
\IEEEcompsocthanksitem {Weizhi Nie, Zichun Zhang, Anan Liu are with the School of Electrical and Information Engineering, Tianjin University, China.\\ E-mail: \{weizhinie, zhangzichun\}@tju.edu.cn, anan0422@gmail.com}
\IEEEcompsocthanksitem {Weijie Wang and Nicu Sebe are with the Department of Information Engineering and Computer Science, University of Trento, Italy. E-mail:\{weijie.wang, niculae.sebe\}@unitn.it. $ ^*$ Corresponding author.}
\IEEEcompsocthanksitem {Bruno Lepri is with Fondazione Bruno Kessler, Italy. E-mail: lepri@fbk.eu}
}
\thanks{Manuscript received April 19, 2005; revised August 26, 2015.}
}

\IEEEtitleabstractindextext{%
\begin{abstract}
Counterfactual medical image generation effectively addresses data scarcity and enhances the interpretability of medical images. However, due to the complex and diverse pathological features of medical images and the imbalanced class distribution in medical data, generating high-quality and diverse medical images from limited data is significantly challenging.
Additionally, to fully leverage the information in limited data, such as anatomical structure information and generate more structurally stable medical images while avoiding distortion or inconsistency. 
In this paper, 
in order to enhance the clinical relevance of generated data and improve the interpretability of the model,
we propose a novel medical image generation framework,
which generates independent pathological and structural features based on causal disentanglement and utilizes text-guided modeling of pathological features to regulate the generation of counterfactual images.
First, we achieve feature separation through causal disentanglement and analyze the interactions between features. Here, we introduce group supervision to ensure the independence of pathological and identity features. 
Second, we leverage a diffusion model guided by pathological findings to model pathological features, enabling the generation of diverse counterfactual images.
Meanwhile, we enhance accuracy by leveraging a large language model to extract lesion severity and location from medical reports. 
Additionally, we improve the performance of the latent diffusion model on long-tailed categories through initial noise optimization.

\end{abstract}

}

\begin{document}

\maketitle
\IEEEdisplaynontitleabstractindextext
\IEEEpeerreviewmaketitle

\section{Introduction}
\begin{figure}[!t]
    \centering
    \includegraphics[width=0.9\linewidth]{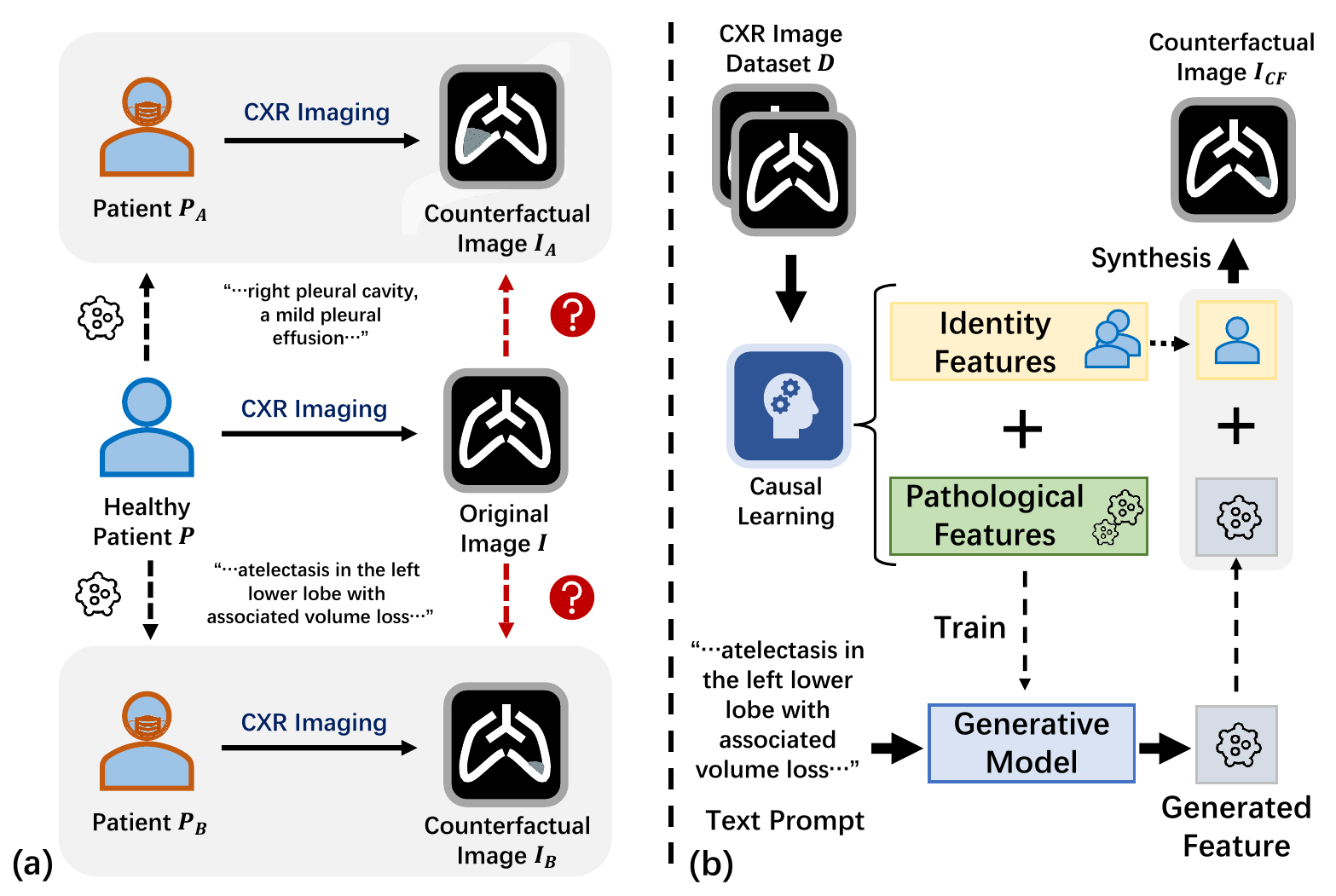}
    \caption{Motivation of our framework. (a) Counterfactual reasoning in the medical domain aims to answer hypothetical questions such as, "What if patient $P$ were affected by disease $A$?" (b) Our proposed method is based on causal disentanglement, which extracts the interactions between pathological and identity features from a large medical dataset $D$. By utilizing text prompts to synthesize specific pathological features, our framework generates diverse counterfactual images $I_{\text{CF}}$, facilitating more interpretable and clinically meaningful medical image synthesis.}
    \label{fig:moti}
\end{figure}
 \IEEEPARstart{I}{n} recent years, deep learning has achieved remarkable progress in the medical domain and significantly improved performance in medical image analysis tasks~\cite{zs1}\cite{seg}. 
 However, challenges persist due to privacy concerns, high data acquisition costs, and the scarcity of rare disease cases, which hinder model training. Specifically, in multi-label tasks like chest X-ray (CXR) analysis, deep learning models are often influenced by data distribution biases and tend to favor common disease classes and overlook rare but clinically important ones \cite{holste2022long}. This limits the model’s generalization ability in low-frequency disease detection \cite{Chlap_2021_06}\cite{mona}. 
 Furthermore, most existing deep learning models lack interpretability, which makes it difficult for clinicians to understand the reasoning behind the model’s predictions, which in turn reduces their trustworthiness and clinical applicability \cite{Choi_2021_10}.

Counterfactual image generation \cite{Goyal_2019_04} provides an effective solution to these challenges. 
In the context of medical image synthesis, counterfactual generation can be understood as addressing hypothetical medical questions \cite{Cohen}.
For example: ``Given that a healthy patient $P$ has a CXR image $I$, what would the CXR image of the same patient (denoted as $P_A$) look like if she/he is diagnosed with disease $A$ (as shown in Fig.~\ref{fig:moti}(a))?”
By synthesizing medical images with specific pathological features, counterfactual generation expands the coverage of data distributions without increasing real-world data collection costs, thus improving model performance on underrepresented disease categories \cite{scorebase}. 
Moreover, counterfactual images allow for fine-grained manipulation of pathological features while preserving individual patient characteristics\cite{rotem2024visual}. 
This enables clinicians to assess model decisions by posing queries such as: ``If this pathological feature were present or absent, would the model still make the same prediction?”
This not only enhances the interpretability of medical image analysis models but also helps clinicians focus on clinically relevant imaging patterns while reducing reliance on spurious correlations, such as the presence of medical devices or acquisition artifacts \cite{xie2020chexplain}. 
Therefore, counterfactual image generation contributes to improving the robustness and clinical reliability of deep learning models in medical imaging.


However, we notice that current medical image generation methods still face the following limitations:
\begin{itemize}
    \item \textbf{Poor Generalizability:} Existing medical image generation methods primarily model simple image class changes, focusing on the transformation between two domains, such as from ``abnormal to healthy'' or ``healthy to abnormal''. These predefined class editing methods make it hard to achieve diverse lesion category generation in multi-label tasks \cite{biomed}.
    \item \textbf{Sample Scarcity:} Due to the high cost, patient privacy concerns, and the inherent rarity of certain cases, medical image data are affected by scarcity and imbalance \cite{holste2022long}. This scarcity of image data makes it difficult for image generation models to train effectively on long-tail categories, leading to inaccurate image synthesis and poor diversity. 
\end{itemize}

\subsection{Motivation}
Current counterfactual generation methods typically disentangle pathological features and structural features of CXR images, so as to extract healthy counterfactual images from the abnormal ones \cite{Wang_2022_01}\cite{ano}\cite{Tang_2020_10}. 
We argue that CXR images are the result of pathological features acting on identity features that are independent of the pathological ones. We hope to explore the mechanisms of pathological features from a large number of medical images. As shown in Fig.~\ref{fig:moti}(b), we aim to flexibly construct pathological feature stimuli to enable the generation of diverse and controllable pathologies under specified structural features, guided by text conditions that describe desired pathological semantics. This approach seeks to fit the real image distribution that is not captured by the original dataset \cite{Ribeiro_2023_01}.
Specifically, we need to address the following fundamental challenges:
\begin{itemize}
    \item How to disentangle and learn the mechanisms of pathological and identity features in medical images is challenging, as these features are often affected by highly coupled factors.
    \item How to improve the image generation quality for long-tail pathological categories. It is essential to mitigate the effects of limited samples and to ensure the image generation model performs well even for rare pathological categories.
\end{itemize}

To resolve these, we start with traditional counterfactual medical image generation methods by pretraining a VAE \cite{VAE} on a large-scale CXR dataset to disentangle pathological features from non-pathological ones. The pathological features are conditioned on an additional constraint provided by a classifier trained on the CXR dataset. Given that unsupervised learning cannot achieve stable feature disentanglement \cite{locatello2019challenging} and that the lack of corresponding data for counterfactual generation makes it challenging to construct direct supervision signals, we employ a group-supervised approach \cite{ge2020zero}. 
Specifically, we swap pathological features extracted from a set of training samples to ensure the accuracy and counterfactual nature of the disentangled features. 
Given that feature vectors are not entirely independent of each other but exhibit certain causal dependencies, we introduce a multi-layer Transformer \cite{trans} to capture the interaction mechanisms among these features. 
After disentanglement, we employ a text-conditioned latent diffusion model (LDM) \cite{rombach2022highresolutionimagesynthesislatent} to characterize pathological features and synthesize diverse counterfactual images by integrating them with the disentangled identity features. Given the high sensitivity of LDM generation to textual guidance, we leverage GPT-4o\cite{gpt4} to extract concise descriptions related to pathological location and severity from medical reports in datasets and eliminate redundant information to provide more precise and effective conditioning. Additionally, by optimizing the initial noise in the inference process \cite{seedselect}, we alleviate the low expressiveness in certain regions of the latent space. This adjustment guides the diffusion model to generate images of long-tail categories.

\subsection{Contribution}
In summary, our contributions are as follows:
\begin{itemize}
    \item We designed a counterfactual medical image synthesis framework that enables diverse and controllable data generation by learning the causal relationships between features in medical images.
    \item We optimized the image synthesis for long-tailed categories in medical data, enhancing the model’s generative performance under class-imbalanced conditions.
    \item  We conducted detailed evaluations of the method in terms of visual quality and pathological features, validating the reliability of our approach.
\end{itemize}


\begin{figure*}
    \centering
    \includegraphics[width=0.9\linewidth]{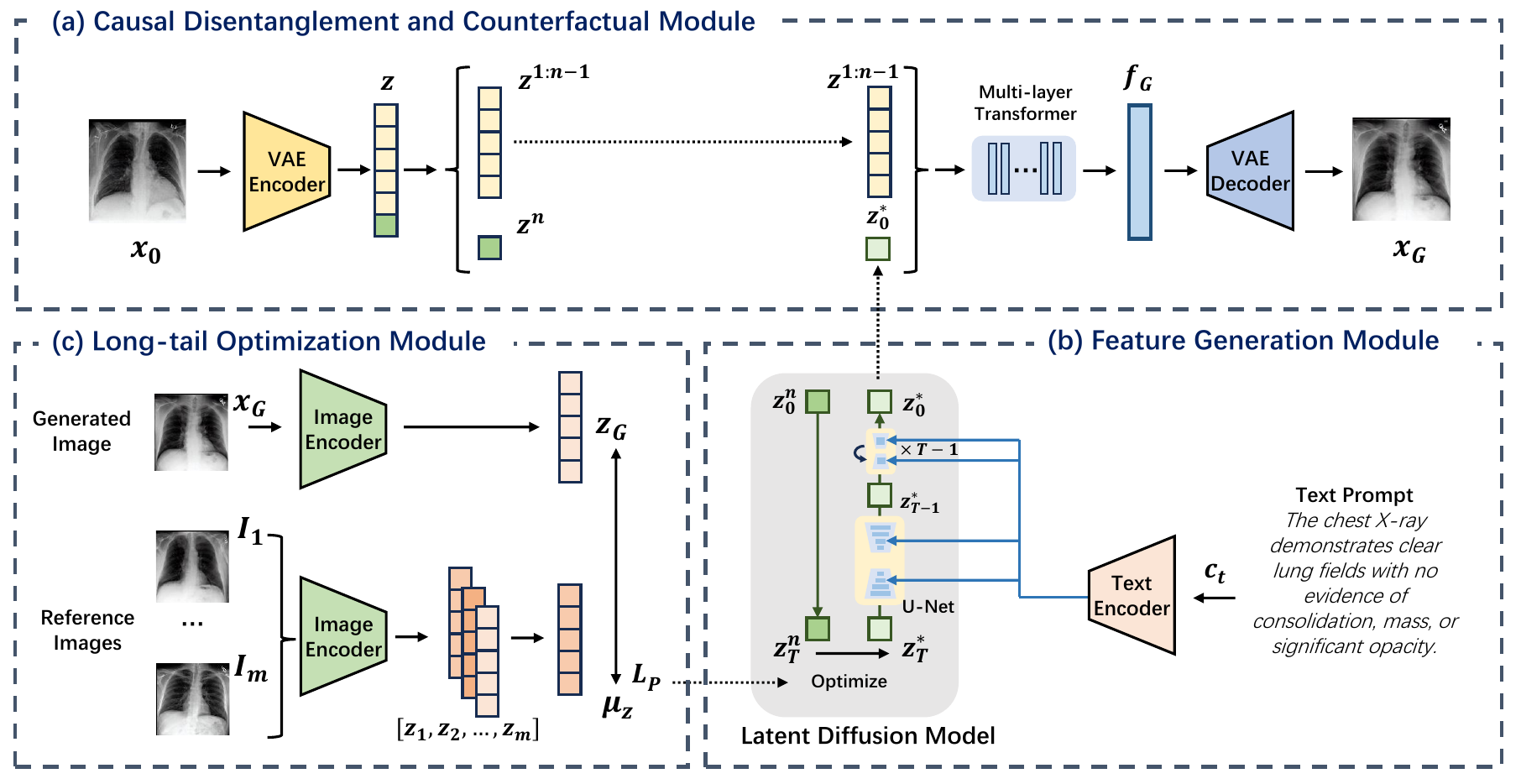}
    \caption{Overview of the framework implementation, which mainly consists of three parts: (a) Causal Disentanglement and Counterfactual Module: The VAE disentangles medical image features, and we employ a multi-layer transformer to learn the interactions between vectors, separating pathological features from structural features in the images. (b) Feature Generation Module: A latent diffusion model is introduced to model the pathological features using text prompts, and by combining the disentangled structural features, it synthesizes counterfactual images. (c) Long-tail Optimization Module: Using $m$ reference images, the initial noise of the generation module is optimized to guide the generated images toward the long-tail categories.}
    \label{fig:method}
\end{figure*}

\section{Related Work}

\subsection{Counterfactual Image Generation}
Counterfactual medical image generation enhances model interpretability by simulating different pathological conditions and domain variations, aiding clinical decision-making. Sanchez et al. \cite{sanchez2022diffusion} and Gu et al.\cite{biomed} utilized diffusion models to generate counterfactual medical images, aiming to support clinical decision-making and enhance model interpretability. Xia et al. \cite{xia2022} and Roschewitz et al. \cite{Roschewitz_2024_09} employed counterfactual image synthesis for data augmentation, thereby improving model robustness. Additionally, Thiagarajan et al.\cite{thiagarajan2022training} and Rotem et al.\cite{rotem2024visual} leveraged counterfactual medical imaging to facilitate clinical decision-making and interpret model predictions. These works improve model robustness, interpretability, and clinical utility.

\subsection{Diffusion Model-Based Medical Image Synthesis}
Diffusion models have become prominent in image generation \cite{hamamci2023generatect}. Ho et al. introduced DDPM\cite{ho2020denoisingdiffusionprobabilisticmodels} for high-quality synthesis via iterative denoising. To enhance efficiency, Rombach et al. proposed Latent Diffusion Models\cite{rombach2022highresolutionimagesynthesislatent}, reducing training costs while maintaining image fidelity.

Due to their significant potential, diffusion models have been widely utilized in the field of medical imaging for image synthesis and data augmentation. These models can substantially enhance data quality and diversity, exemplified by applications in magnetic resonance imaging (MRI) demonstrated by Pan et al.\cite{pan2021disease}  and Pinaya et al. \cite{pinya} and optical coherence tomography (OCT) by Hu et al. \cite{hu2022unsupervised}.
Researchers, including Luo et al.\cite{luo2024measurement}, Ali et al.\cite{ali2022spot}, and Packhäuser et al.\cite{packhauser2023generation} employed class-conditional diffusion models to specifically target particular diseases or lesion types, generating supplementary medical images to augment datasets, thereby improving performance in diagnostic and classification tasks. Additionally, Chambon et al. \cite{roentgen} adapted pre-trained diffusion models to the medical domain to generate synthetic chest X-ray images characterized by specific abnormalities and attributes. This approach facilitates customized image generation guided by natural language descriptions.
 Rouzrokh et al.\cite{rouz} and Wolleb et al.\cite{woll} utilized the inpainting capabilities of diffusion models to remove or precisely localize abnormal regions in medical images, thereby supporting clinical diagnostics or image reconstruction tasks. 
 
 Our proposed approach adopts a novel perspective by focusing the model on pathological feature modeling, enabling diverse counterfactual image synthesis while preserving structural information.

\subsection{Disentangled representation learning}
Disentangled representation learning separates meaningful factors in data, improving interpretability and generalization. In medical image analysis, it enables segmentation, synthesis, and privacy preservation by isolating domain-invariant and task-specific features \cite{sur}. Xie et al. \cite{Xie}, Chartsias et al. \cite{Chart}, and Liu et al. \cite{Liu_2021_01} demonstrated its efficacy in domain-adaptive segmentation. Tang et al. \cite{Tang_2020_10} and Reaungamornrat et al. \cite{cross} applied disentanglement to disease detection and cross-modal synthesis. Han et al. \cite{Han} suppressed rib structures in CXR images for improved pulmonary disease diagnosis. Xia et al. \cite{Xia_2020_06} generated pseudo-healthy images by disentangling pathological and healthy information for anomaly detection. Montenegro et al. \cite{ano} proposed identity-preserving disentanglement for privacy-preserving medical image synthesis. CausalVAE \cite{causalvae} integrates causal inference with VAE to achieve interpretable disentanglement and counterfactual synthesis. While disentanglement learning has shown promise, the complex interactions between medical image features make direct modeling challenging, increasing the difficulty of achieving effective disentanglement. Our method learns the latent effects of pathology through large-scale data, making it better suited for complex tasks such as medical image generation.

\subsection{Long-tail Image Generation}

While diffusion models excel at high-fidelity image synthesis, they face notable challenges in generating rare concepts and require long training times. Qin et al. \cite{clsimb} highlight the issue of "class forgetting" and "head-class bias" in diffusion models, where imbalanced training data leads to significantly degraded generation quality for tail classes.
Samuel et al. \cite{seedselect} observed that diffusion models only cover a limited portion of the input space during training, leading to failures when generating underrepresented concepts. To address this, they proposed SeedSelect \cite{seedselect}, which optimizes noise selection in the latent space, significantly improving long-tail categories generation without additional model training.

In medical imaging, dataset imbalance arises due to high acquisition costs and the rarity of certain conditions. While prior methods \cite{longtail1}\cite{longtail2}\cite{longtail4} attempt to mitigate these challenges, they often lack diversity and efficiency. Inspired by this, we propose a training-free optimization approach to enhance generation diversity and ensure reliable synthesis of long-tail category medical images.

\section{Method}
In this section, we detail our method. Fig.~\ref{fig:method} illustrates our framework, which includes three key components: causal disentanglement and counterfactual module, feature generation module, and long-tail optimization module.

\subsection{Counterfactual Generation Framework}
To enable counterfactual medical image synthesis, we first disentangle chest X-ray image $x_0$ into identity-related and pathology-related latent features with $\mathcal{E}_\text{VAE}$. Specifically, we represent the latent code as $z = \{z^{1:n-1}, z^n\}$, where $z^{1:n-1}$ encodes identity features such as anatomical structure and patient-specific characteristics, and $z^n$ captures pathology-related factors.

Unlike conventional approaches that assume independence among latent factors, we posit that these components exhibit causal dependencies—for instance, pathological features may affect anatomical appearance. To capture these interactions, we introduce a causal disentanglement mechanism that models how $z^n$ influences the overall image representation.

We assume that the final latent vector is determined by a causal process:
\begin{equation}
z = g(z^{1:n-1}, z^n, \epsilon)
\end{equation}
where $g(\cdot)$ denotes the underlying causal mechanism, and $\epsilon \sim \mathcal{N}(0, I)$ represents stochastic noise independent of semantic content.

Since the true function $g(\cdot)$ is unknown, we approximate it using a Transformer network with self-attention:
\begin{equation}
f = \mathcal{T}(z)
\end{equation}
Here, $f$ is the causally modulated latent representation. The Transformer learns the internal interactions between identity and pathological features, allowing the model to generate anatomically consistent and pathologically accurate counterfactual images.

We generate a pathology feature vector $z^n$ using a text-guided latent feature generator, and refine it through long-tail optimization, and an optimized initial noise seed $z^*_T$ is denoised to obtain the final vector $z^*_0$. This optimized feature is then combined with the identity vector $z^{1:n-1}$, and the resulting latent representation is processed by the Transformer module as:
\begin{equation}
f_G = \mathcal{T}(g(z^{1:n-1}, z^*_0, \epsilon))
\end{equation}

Finally, the causally modulated representation $f_G$ is decoded by the VAE decoder $\mathcal{D}_{\text{VAE}}$ to synthesize the medical image $x_G$.

\subsubsection{Causal Disentanglement Module}

This module focuses on causally disentangling identity and pathological features to support accurate image reconstruction and controllable counterfactual generation. To ensure stable and meaningful disentanglement, it is essential to introduce effective supervision signals.

\begin{figure}[t]
    \centering
    \includegraphics[width=1\linewidth]{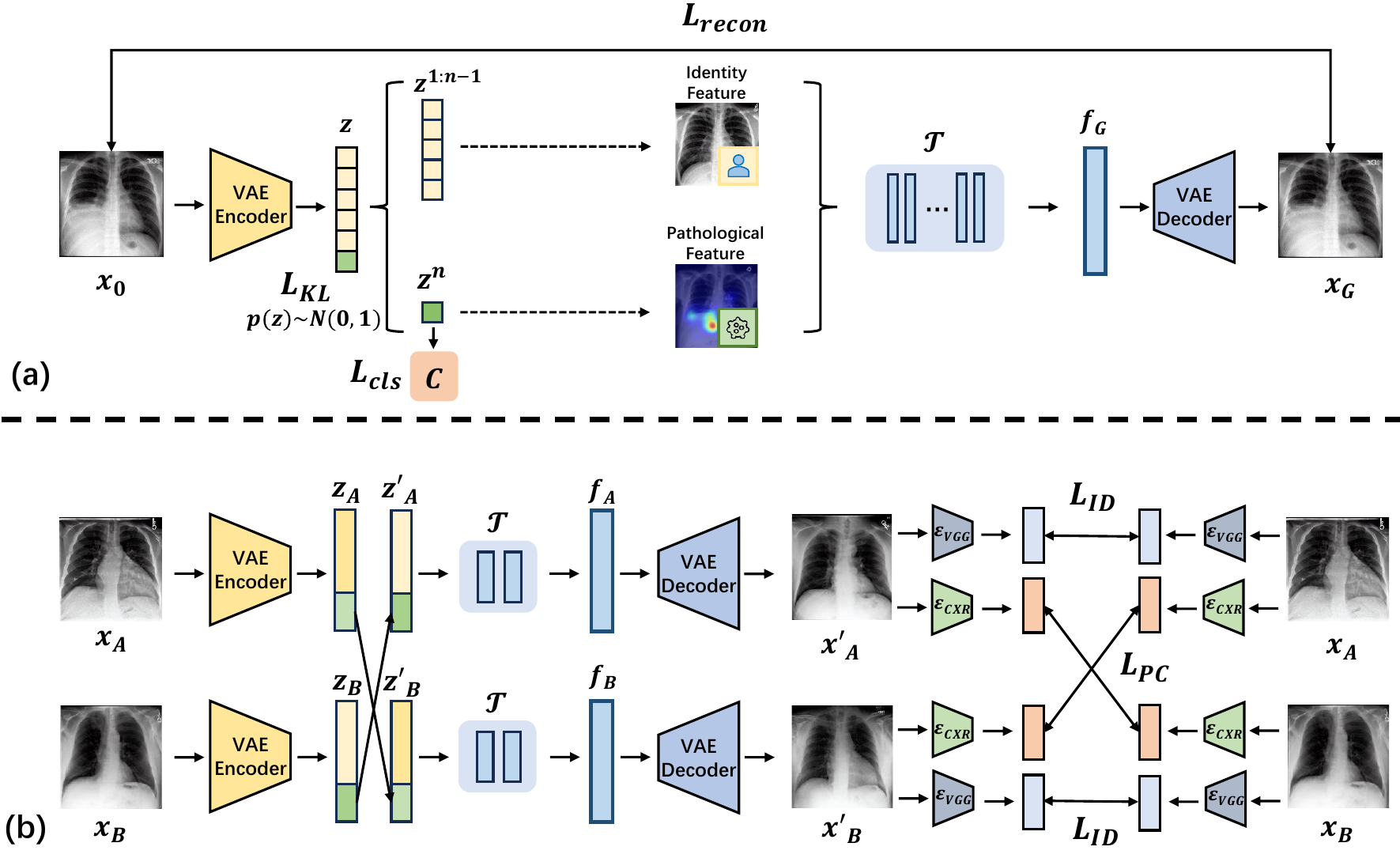}
    \caption{Training process of the causal disentanglement and counterfactual generation module. (a) A medical image $x_0$ is encoded by a VAE, where $z^n$ represents pathological features and $z^{1:n-1}$ represents identity features. A multi-layer Transformer $T$ models their interactions, generating transformed features $f$, which are decoded into $x_G$. (b) To ensure counterfactual consistency, we use a pre-trained VGG-16 for identity features and CXR-CLIP for pathology features. A group supervision mechanism swaps features to provide supervision signals, ensuring accurate disentanglement.}
    \label{fig:VAE}
\end{figure}

To construct such signals, we adopt a group-supervised learning approach \cite{ge2020zero} to train the module. As shown in Fig.~\ref{fig:VAE}(b), we first select two medical images, $x_A$ and $x_B$, from the training dataset and extract their corresponding latent representations, $z_A$ and $z_B$. To enforce disentanglement, we exchange their pathological features $z^n$, creating new latent representations, $z'_A$ and $z'_B$. These modified feature representations are then processed by the Transformer to capture nonlinear interactions and subsequently decoded to generate new images, $x'_A$ and $x'_B$. Through this process, the generated image $x'_A$ retains the identity information of $x_A$ while incorporating the pathological characteristics of $x_B$, ensuring a structured and controlled counterfactual generation.

To train the framework, we employ a combination of loss functions:

\begin{itemize}
    \item \textbf{Reconstruction Loss} $L_{\text{recon}}$: Ensures that the decoder reconstructs the input image accurately by minimizing both pixel-wise differences and structural inconsistencies. 
    \begin{equation}
    L_{\text{recon}} = \mathbb{E}_{q(z|x)} \left[ \| x - \hat{x}(z) \|^2 + \lambda (1 - SSIM(x, \hat{x}(z))) \right],
    \end{equation}
    where $\lambda$ is a weighting factor balancing pixel-wise accuracy and perceptual similarity. SSIM helps preserve fine-grained details and structural integrity during reconstruction.

    \item \textbf{KL Divergence Loss} $L_{KL}$: Regularizes the latent space by encouraging the learned distribution $q(z|x)$ to approximate the standard normal prior $p(z) = \mathcal{N}(0, I)$. This constraint ensures that the latent space remains smooth and well-structured, preventing overfitting.
    \begin{equation}
    L_\text{KL} = D_{KL}[ q(z|x) \| p(z) ].
    \end{equation}
   
    \item \textbf{Classification Loss} $L_{cls}$: Ensures that $z^n$ correctly encodes pathology-related information by enforcing a classification objective. It uses cross-entropy loss to train a classifier that predicts the correct pathology label based on $z^n$.
    \begin{equation}
    L_\text{cls} = - \mathbb{E}_{q(z|x)} \left[ \sum_{i=1}^{n_{cls}} y_i \log P(y_i | z^n) \right],
    \end{equation}
    where $y_i$ is the one-hot encoded ground truth label, $n_{cls}$ represents the total number of pathology categories in the dataset, and $P(y_i | z^n)$ is the predicted probability of the pathology class.

    \item \textbf{Identity Consistency Loss} $L_\text{ID}$: Ensures that patient identity remains unchanged despite pathological interventions. This loss minimizes the Euclidean distance in the identity feature space extracted by a pre-trained VGG-16 \cite{VGG} model:
    \begin{equation}
    L_\text{ID} = \sum_{i \in \{A, B\}} \left\| \mathcal{E}_{\text{VGG}}(x_i) - \mathcal{E}_{\text{VGG}}(x'_i) \right\|^2,
    \end{equation}
    where $\mathcal{E}_{\text{VGG}}(\cdot)$ is the identity feature extractor.

    \item \textbf{Pathological Consistency Loss} $L_\text{PC}$: Ensures that pathology-related features remain consistent after counterfactual transformations. This loss minimizes the Euclidean distance between the original and transformed pathology-related features extracted by a CXR-CLIP encoder:
    \begin{equation}
    L_\text{PC} = \sum_{i \in \{A, B\}} \left\| \mathcal{E}_{\text{CXR}}(x_i) - \mathcal{E}_{\text{CXR}}(x'_i) \right\|^2,
    \end{equation}
    where $\mathcal{E}_{\text{CXR}}(\cdot)$ is the 
encoder to extract pathology features.
\end{itemize}

The final objective function consists of multiple loss components to ensure effective disentanglement and counterfactual generation. We define the VAE loss as:

\begin{equation}
L_{\text{VAE}} = L_{\text{recon}} + \alpha L_{\text{KL}} + \beta L_{\text{cls}}
\end{equation}

The final objective function is formulated as:
\begin{equation}
L_{\text{total}} = L_{\text{VAE}} + \gamma L_{\text{ID}} + \delta L_{\text{PC}}
\end{equation}

where $\alpha, \beta, \gamma, \delta$ are hyperparameters that control the relative contributions of each loss term.

To reduce the training cost, we apply $L_{\text{ID}}$ and $L_{\text{PC}}$ as supervision signals only at specific training steps, while setting $ \gamma $ and $ \delta $ to zero for the remaining steps.

By leveraging causal disentanglement, our framework establishes a structured latent space where pathological factors can be selectively manipulated, thereby enabling controllable counterfactual medical image synthesis. This allows us to generate realistic medical images with altered pathological conditions while preserving patient identity.

\begin{figure}[t]
    \centering
    \includegraphics[width=1\linewidth]{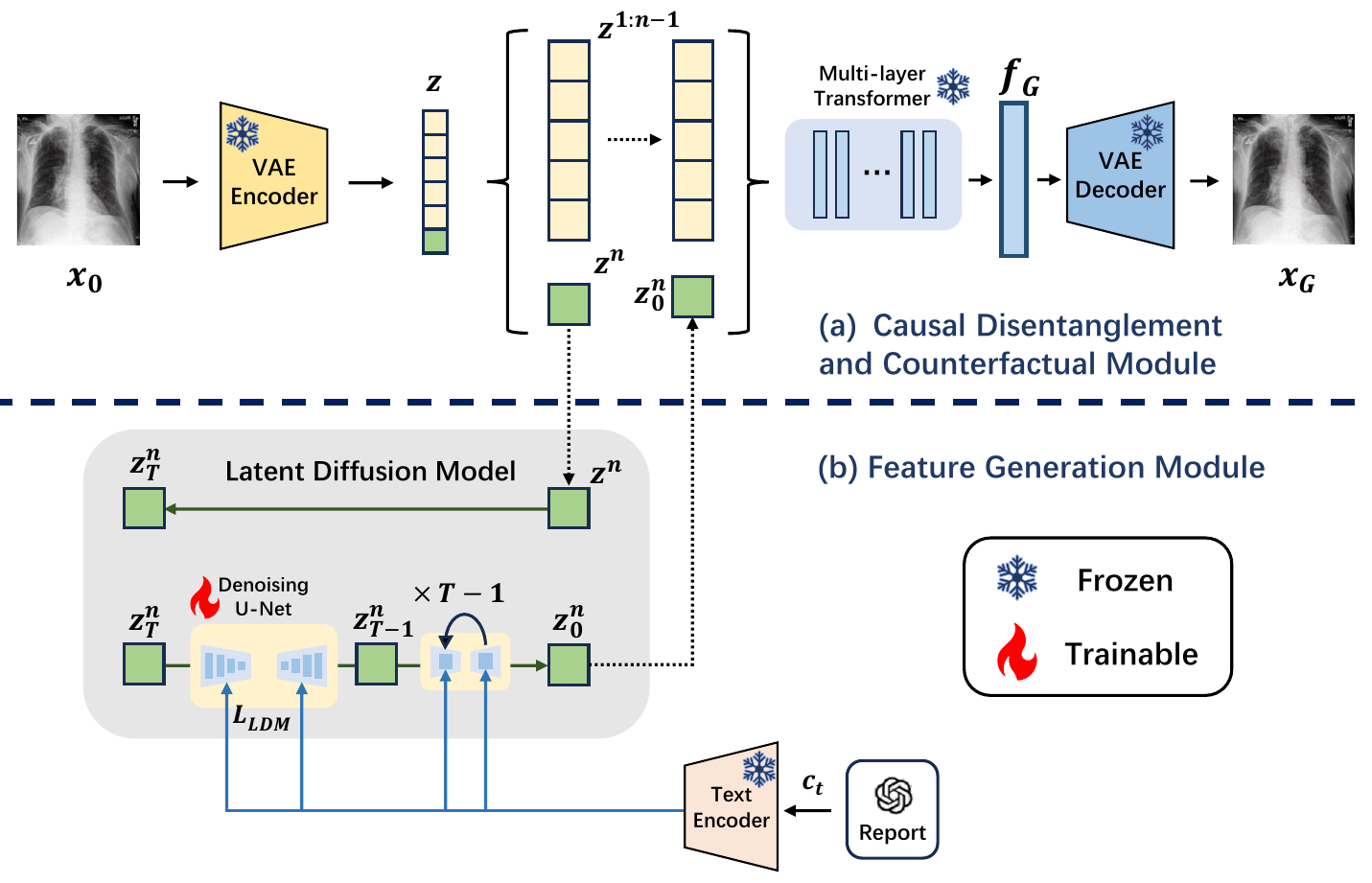}
    \caption{Training process of the feature generation module. After the VAE encoder disentangles the image features, a text-conditioned diffusion model generates the pathological features $z_{0}^{n}$. The decoder then integrates the identity and pathological features to synthesize the final image.}
    \label{fig:CFG}
\end{figure}

\subsubsection{Feature Generation Module}

Conventional counterfactual image-generation methods primarily rely on adversarial learning, which often lacks flexible conditional control. To address this limitation, we introduce a text-conditioned latent diffusion model for generating pathological information. Unlike previous approaches, our method operates in the latent space disentangled by the VAE, allowing for more diverse and controllable image generation. Additionally, by optimizing the noise space of the latent diffusion model, we can effectively mitigate the challenges associated with synthesizing images for long-tail categories.


The latent diffusion model, proposed by Rombach et al. based on DDPM \cite{ho2020denoisingdiffusionprobabilisticmodels}, introduces a more efficient approach to diffusion-based image synthesis. Unlike standard diffusion models that operate directly in the high-dimensional pixel space ($x_0 \in \mathbb{R}^n$), LDMs perform diffusion in a lower-dimensional learned latent space $z_0$, which is obtained by encoding the input image $x_0$ using a VAE encoder.

In the latent space, the forward diffusion process gradually corrupts $z_0$ into a Gaussian noise distribution. It is defined as:
\begin{equation}
q(z_t | z_{t-1}) = \mathcal{N}(z_t; \sqrt{\alpha_t} z_{t-1}, (1 - \alpha_t) I),
\end{equation}
where $z_t$ represents the latent representation at time step $t$, and $\alpha_t = 1 - \beta_t$ is the noise scaling factor.

The reverse diffusion process iteratively removes noise from $z_t$, reconstructing the original latent representation $z_0$. The generative model defines this process as:
\begin{equation}
p_\theta(z_{t-1} | z_t) = \mathcal{N}(z_{t-1}; \mu_\theta(z_t, t), \sigma_t^2 I),
\end{equation}
where $\mu_\theta(z_t, t)$ is the predicted mean of the denoised distribution, parameterized by a neural network.
Instead of directly predicting $\mu_\theta(z_t, t)$, as in standard DDPMs, LDM learns to predict the noise $\epsilon$ added during the forward process. The denoised mean $\mu_\theta(z_t, t)$ can be computed as:
\begin{equation}
\mu_\theta(z_t, t) = \frac{1}{\sqrt{\alpha_t}} \left( z_t - \frac{1 - \alpha_t}{\sqrt{1 - \bar{\alpha}_t}} \epsilon_\theta(z_t, t) \right),
\end{equation}
where $\bar{\alpha}_t = \prod_{s=1}^{t} \alpha_s$ represents the cumulative noise schedule.

Following the DDPM framework, LDM is trained by minimizing the following noise prediction loss:
\begin{equation}
L_{\text{LDM}} = \mathbb{E}_{t, z_0, \epsilon} \left[ \|\epsilon - \epsilon_\theta(z_t, t)\|^2 \right].
\end{equation}

\begin{figure}[t]
    \centering
    \includegraphics[width=1\linewidth]{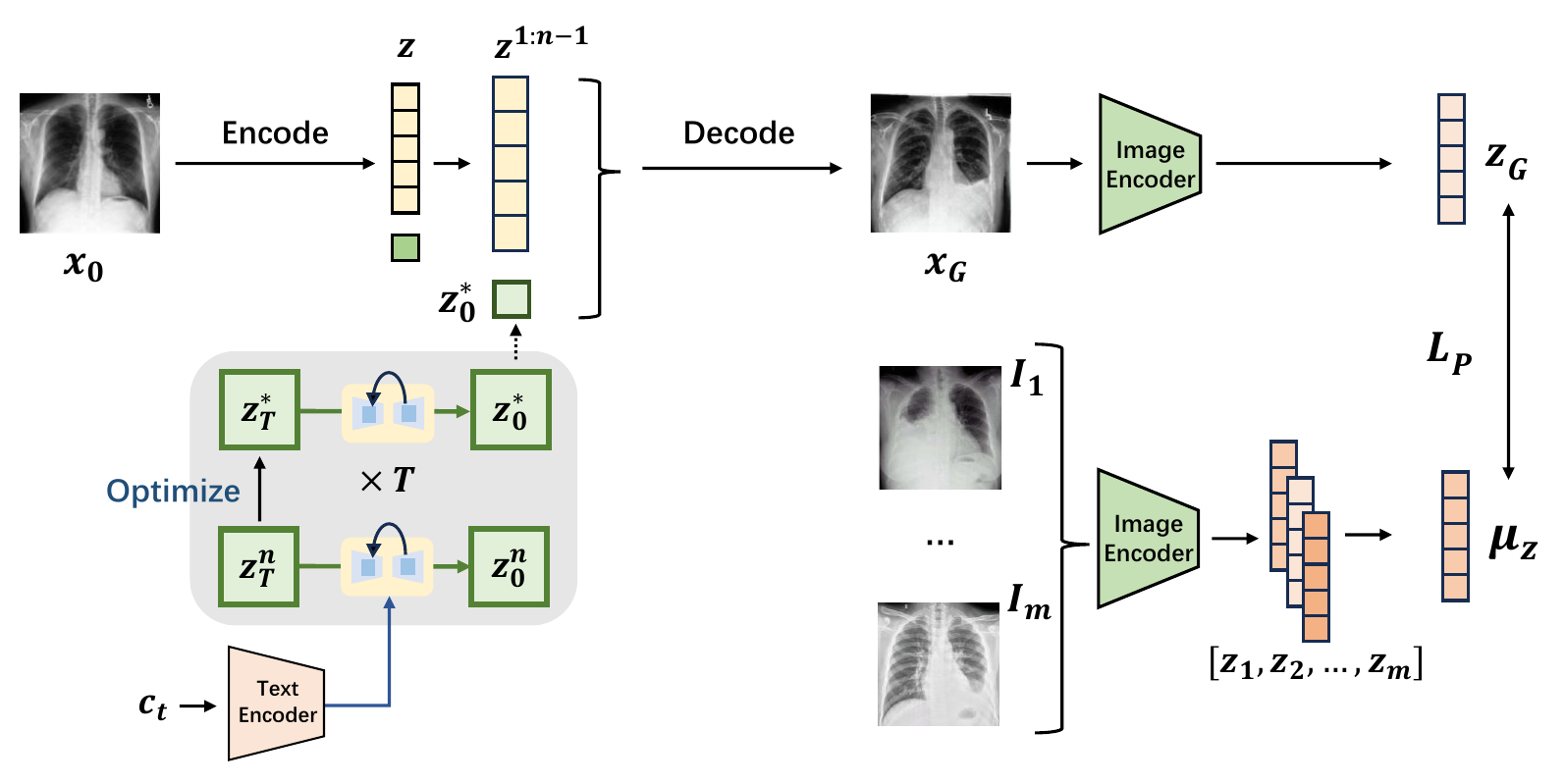}
    \caption{Long-tail optimization. By optimizing the initial noise $z^n_T$ for pathology feature generation, the semantic distance between the generated image $x_G$ and the reference images from the long-tail category is minimized.}
    \label{fig:fs}
\end{figure}

Unlike conventional generative methods, our diffusion model does not model the entire image space. Instead, it focuses solely on generating pathological features $z^n$, as formulated by:
\begin{equation}
    p_{\theta_n} (z^n_0) = \int p_{\theta_n}(z^n_T) \prod_{t=1}^{T} p_{\theta_n}(z^n_{t-1} | z^n_t) \, dz^n_{1:T}, \quad z^n_T \sim \mathcal{N}(0, I).
\end{equation}
The training objective of the diffusion model is to reconstruct $z^n_0$ from noisy samples $z^n_T$. The corresponding loss function is defined as:
\begin{equation}
    L_{G} = \mathbb{E}_{z^n_0, t, c_t, \epsilon \sim \mathcal{N}(0,1)} \left[ \left\| \epsilon - \epsilon_{\theta}(z^n_T, t, c_t) \right\|_2^2 \right].
\end{equation}

This loss function ensures that the model learns to progressively denoise the pathological feature vector $z^n$ using text-guided conditioning.

The combination of VAE-based feature disentanglement and diffusion-based generative modeling ensures that counterfactual images maintain anatomical consistency while effectively capturing diverse pathological variations.

\subsection{Long-tail Optimization}
Text guided diffusion models often face challenges in long-tail category generation tasks, particularly in accurately generating underrepresented concepts. According to Samuel et al.~\cite{seedselect}, this issue primarily arises due to the underrepresentation of rare concepts in the dataset, leading to incomplete coverage of the noise space. Since the training process does not adequately span the full noise distribution of rare categories, the model may misinterpret random noise as out-of-distribution samples, resulting in inaccurate image synthesis. 

In medical image generation, we leverage a small set of real medical images ($I_1, I_2, \dots, I_m$) from specific categories to optimize the initial noise input $z^n_T$. This enables the model to generate medical images that better align with pathological characteristics.

As illustrated in Fig.~\ref{fig:fs}, pathological consistency is enforced by measuring the semantic similarity between the generated medical image $I_G$ and a reference set of real medical images $I_1, \dots, I_m$. We employ a pre-trained image encoder to extract pathological feature vectors $z_1, \dots, z_m$ from these reference images. The centroid feature vector is computed as:
\begin{equation}
    \mu_z = \frac{1}{m} \sum_{i=1}^{m} z_i.
\end{equation}

Similarly, we encode the generated medical image $I_G$ to obtain its feature representation $z_G$. The pathological semantic loss is then formulated as:
\begin{equation}
    L_{\text{P}} = \| \mu_z - z_G \|^2,
\end{equation}
where $\| \cdot \|^2$ represents the Euclidean distance in the feature space between the centroid feature vector $\mu_z$ and the generated image feature vector $z_G$. This loss function ensures that the generated image accurately preserves the pathological characteristics present in the reference images.

During optimization, the total loss $L_{\text{P}}$ is backpropagated to iteratively refine the initial noise input $z^n_T$, resulting in an optimized one $z^*_T$. Notably, this approach does not modify the parameters of the pre-trained diffusion model, thereby preserving computational efficiency and ensuring model stability, while achieving more precise long-tail category medical image synthesis.

\subsection{Overall Pipeline}
\begin{algorithm}[t]
\caption{Optimized Training Procedure for Causal Disentanglement and Counterfactual Generation}
\label{alg:optimized}
\begin{algorithmic}[1]
\Require 
  Dataset $D$ of images $(x)$ with attributes, labels $y$, and text reports $c_t$, 
  learning rates $lr_1$, $lr_2$, 
  weights $\alpha$, $\beta$, $\gamma$, $\delta$
\Ensure 
  Parameters $\theta_{\mathcal{E}_{\text{VAE}}}, \theta_{\mathcal{D}_{\text{VAE}}}, \theta_{\mathcal{T}}, \theta_{\text{UNet}}, \theta_{\mathcal{E}_{\text{Text}}}$

\textbf{Stage 1: Train VAE and Transformer}
\State Initialize $\theta_{\mathcal{E}_{\text{VAE}}}, \theta_{\mathcal{D}_{\text{VAE}}}, \theta_{\mathcal{T}}$
\State Initialize optimizer with learning rate $lr_1$
\For{each training step $t$ on $D$}
    \State $x \gets$ sample an image from $D$
    \State $z \gets \mathcal{E}_{\text{VAE}}(x; \theta_{\mathcal{E}_{\text{VAE}}})$
    \If{$t \mod 5 = 0$} 
        \State $x_A, x_B \gets$ sample a pair of images from $D$ 
        \State $z_A = \mathcal{E}_{\text{VAE}}(x_A; \theta_{\mathcal{E}_{\text{VAE}}}), z_B = \mathcal{E}_{\text{VAE}}(x_B; \theta_{\mathcal{E}_{\text{VAE}}})$ 
        \State $f_A \gets \mathcal{T}(z_A; \theta_{\mathcal{T}}),f_B \gets \mathcal{T}(z_B; \theta_{\mathcal{T}})$  
        \State $z'_A = \{ z_A^{1:n-1}, z_B^n \},z'_B = \{ z_B^{1:n-1}, z_A^n \}$  
        \State $x'_A = \mathcal{D}_{\text{VAE}}(z'_A; \theta_{\mathcal{D}_{\text{VAE}}}),x'_B = \mathcal{D}_{\text{VAE}}(z'_B; \theta_{\mathcal{D}_{\text{VAE}}})$
        
        \State 
        $L_{\text{ID}} = \sum_{i \in \{A, B\}} \left\| \mathcal{E}_{\text{VGG}}(x_i) - \mathcal{E}_{\text{VGG}}(x'_i) \right\|^2$   
        \State
        $L_{\text{PC}} = \sum_{i \in \{A, B\}} \left\| \mathcal{E}_{\text{CXR}}(x_i) - \mathcal{E}_{\text{CXR}}(x'_{i}) \right\|^2$

    \Else
        \State $L_{\text{ID}} = 0$, $L_{\text{PC}} = 0$ 
    \EndIf
    \State $L_{\text{recon}} = \| x - \hat{x}(z) \|^2 + \lambda (1 - SSIM(x, \hat{x}(z)))$
    \State 
    $L_{\text{KL}} = D_{KL} \left( q(z|x) \parallel p(z) \right)$
    \State 
    $L_{\text{cls}} = H(y, P(y | z^n))$

    \State 
    $L_{\text{VAE}} = L_{\text{recon}} + \alpha L_{\text{KL}} + \beta L_{\text{cls}}$
    
    \State 
    $L_{\text{total}} = L_{\text{VAE}} + \gamma L_{\text{ID}} + \delta L_{\text{PC}}$
    
    \State $L_{\text{total}}$, update $\theta_{\mathcal{E}_{\text{VAE}}}, \theta_{\mathcal{D}_{\text{VAE}}}, \theta_{\mathcal{T}}$
\EndFor

\textbf{Stage 2: Train Latent Diffusion Model on $z^n$}
\State Initialize $\theta_{\text{UNet}}$
\State Initialize optimizer with learning rate $lr_2$
\For{each training step on $D$}
    \State $(x, c_t) \gets$ sample batch of image-text pairs from $D$
    \State $z = \{z^{1:n-1}, z^n\} \gets \mathcal{E}_{\text{VAE}}(x; \theta_{\mathcal{E}_{\text{VAE}}})$
    \State $t \gets$ sample diffusion timestep
    \State $\epsilon \sim \mathcal{N}(0,I)$
    \State $z^n_T = \sqrt{\bar{\alpha}_t} z^n + \sqrt{1 - \bar{\alpha}_t} \epsilon$
    \State $e_t \gets \mathcal{E}_{\text{Text}}(c_t; \theta_{\mathcal{E}_{\text{Text}}})$
    \State $\hat{\epsilon} = \text{UNet}(z^n_T, t, e_t; \theta_{\text{UNet}})$
    \State $L_G = \mathbb{E} \left[ \| \epsilon - \hat{\epsilon} \|^2 \right]$
    \State Backpropagate $L_G$, update $\theta_{\text{UNet}}$
\EndFor

\State \textbf{return} $\theta_{\mathcal{E}_{\text{VAE}}}, \theta_{\mathcal{D}_{\text{VAE}}}, \theta_{\mathcal{T}}, \theta_{\text{UNet}}$
\end{algorithmic}
\end{algorithm}

\begin{algorithm}[t]
\caption{Long-tail Optimization Inference for Image Generation}
\label{alg:fs_opt}
\begin{algorithmic}[1]
\Require 
  Pre-trained counterfactual generation model $M$, 
  CXR-CLIP encoder $\mathcal{E}_{\text{CXR}}$, 
  text prompt $c_t$, 
  reference images $\{I_1, I_2, ..., I_m\}$, 
  input image $x_0$, 
  number of iterations $N$
\Ensure 
  Optimized seed $z^*_T$, optimized generated image $I_G$
  
\State Initialize latent noise:$z^n_T \sim \mathcal{N}(0, I)$  

\State $z_1, z_2, ..., z_m \gets \mathcal{E}_{\text{CXR}}(I_1, I_2, ..., I_m; \theta_{\mathcal{E}_{\text{CXR}}})$
\State $\mu_z = \frac{1}{m} \sum_{i=1}^{m} z_i$
\State $z=\{z^{1:n-1},z^n\} \gets \mathcal{E}_{\text{VAE}}(x_0; \theta_{\mathcal{E}_{\text{VAE}}})$

\For{$t = 1$ to $N$}
    \State $\hat{\epsilon} = M(z^n_T, c_t)$
    \State $z^n_0 \gets z^n_T - \hat{\epsilon}$      
    \State $f \gets \mathcal{T}(\{z^{1:n-1},z^n_0\}; \theta_{\mathcal{T}})$
    \State $I_G \gets \mathcal{D}_{\text{VAE}}(f;\theta_{\mathcal{D}_{\text{VAE}}})$
    
    \State $z_G \gets \mathcal{E}_{\text{CXR}}(I_G)$
    \State $L_{\text{P}} = \| \mu_z - z_G \|_2$
    \State Backpropagate $L_{\text{P}}$, update $z^n_T$ 
\EndFor

\State \textbf{return} Optimized seed $z^*_T$, final image $I_G$
\end{algorithmic}
\end{algorithm}

To train the VAE-based causal disentanglement architecture, we adopt the loss function defined in Section 3.1. This loss function jointly optimizes both the VAE encoder-decoder and the multi-layer Transformer, ensuring that the disentangled vector $z^n$ effectively captures pathological features while maintaining independence from non-pathological factors.

Once the causal disentanglement module is trained, we introduce a text-conditioned latent diffusion model to further refine the feature vectors, as illustrated in Fig.~\ref{fig:CFG}. The medical report $c_t$ is processed by a text encoder, generating a textual embedding $e_t$ that guides the synthesis of pathological features. Since diffusion models are highly sensitive to text prompts, we utilize GPT-4o to process medical reports, retaining only lesion-related descriptions to enhance alignment between textual information and the generated pathological features.

The training process for both the causal disentanglement module and the counterfactual generation module is outlined in Algorithm 1. During inference, we optimize the noise space using the pathological semantic loss $L_\text{P}$ (introduced in Section 3.2), guiding the model toward more accurate long-tail category medical image synthesis. The inference procedure is detailed in Algorithm 2.

\section{Experiments and Results}
\subsection{Experimental Setup}

\textbf{Datasets.} We conduct experiments on three widely-used chest X-ray datasets: CheXpert~\cite{chexpert}, MIMIC-CXR~\cite{MIMICCXR}, and IU-Xray~\cite{OpenI}. 

\textbf{CheXpert} is a large-scale chest X-ray dataset containing 14 pathology classes. Each class is annotated with one of four possible labels: \textit{positive}, \textit{negative}, \textit{uncertain}, or \textit{none} (not mentioned). The labels are extracted from the impression sections of radiology reports using an automated labeler, but the full-text reports are not available.
\textbf{MIMIC-CXR} is a large-scale dataset that includes both chest X-ray images and free-text radiology reports. It consists of 473,057 images paired with 206,563 radiology reports from 63,478 unique patients.
\textbf{IU-Xray} is a smaller dataset containing 3,684 image-report pairs. It is split into 2,912 samples for training and 772 for testing.

\begin{figure}[t]
    \centering
    \includegraphics[width=1\linewidth]{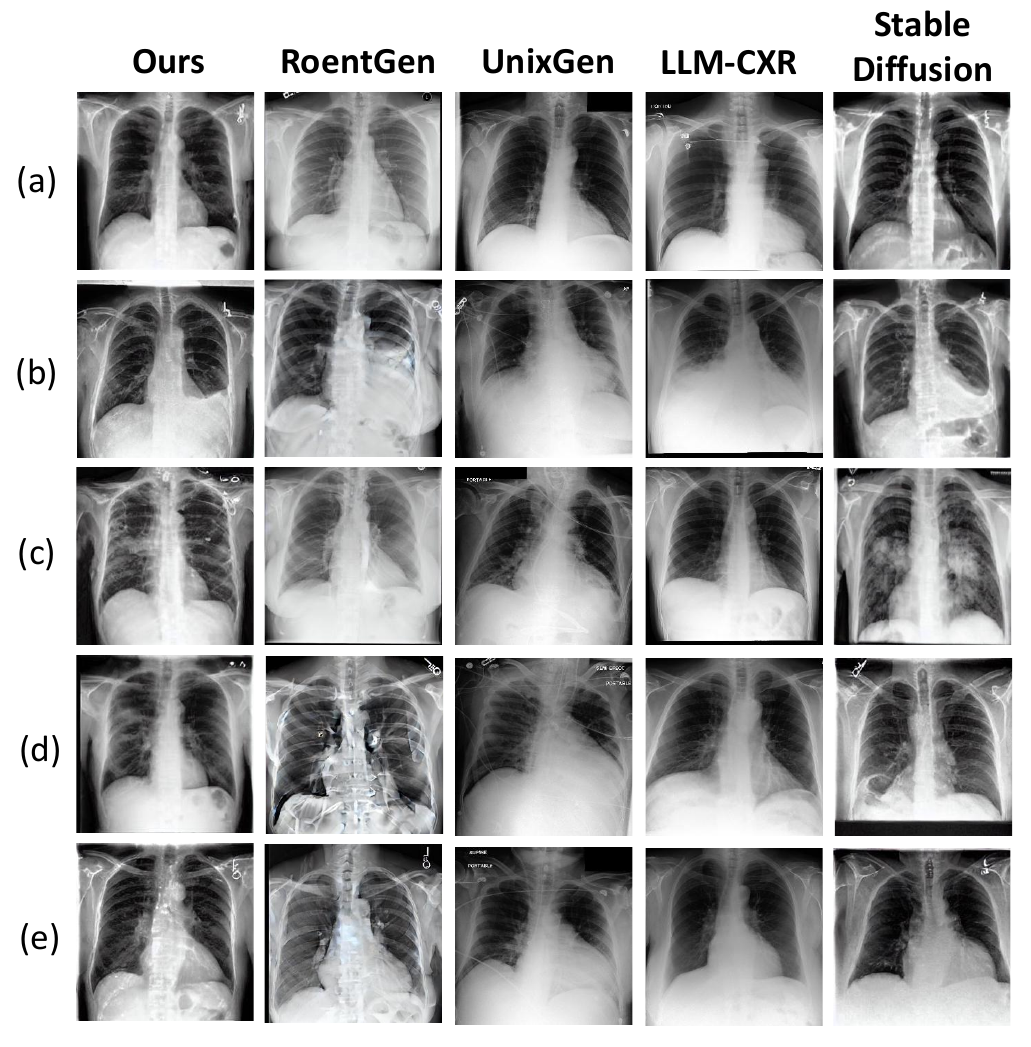}
    \caption{Visual comparison of our model with SOTA medical text-to-image generation models. The five rows correspond to different conditions: (a) no finding, (b) pleural effusion,(c) consolidation, (d) atelectasis , and (e) cardiomegaly. Our method demonstrates superior anatomical consistency and pathology accuracy compared to these models.}
    \label{fig:sota}
\end{figure}

\textbf{Implementation Details.}
In the training stage of the causal disentanglement module, we train the VAE for 100 epochs on the CheXpert and MIMIC-CXR datasets. The VAE encoder compresses the input images of size $256 \times 256$ into a latent representation $z$ of size $(n, 64, 64)$, where $n$ represents the number of disentangled feature channels. The hyperparameters are set as $\alpha = 0.1$, $\beta = 5.0$, and $\lambda = 0.01$ to balance KL divergence, classification loss, and structural similarity constraints, while $\gamma = 1.0$ and $\delta = 2.5$ control the strength of identity and pathology consistency constraints. The training is conducted on 8 NVIDIA A100 GPUs, with a batch size of 128, a learning rate of 1e-4, and an image resolution of 256 × 256, taking approximately 35 hours to complete.
\begin{table}[t]
\centering
\caption{Comparison of text-to-image generation performance on the MIMIC-CXR and the IU-Xray datasets.}
\begin{tabular}{ccccccc}
\toprule
\multirow{2}{*}{\textbf{Method}} & \multicolumn{2}{c}{\textbf{MIMIC-CXR}} & & \multicolumn{2}{c}{\textbf{IU-Xray}} \\ \cmidrule(lr){2-3} \cmidrule(lr){5-6}
                 & \textbf{FID↓} & \textbf{NIQE↓} & & \textbf{FID↓} & \textbf{NIQE↓} \\ \midrule
Stable Diffusion & 21.9198        & 7.7448         & & 21.3165 & 8.1452  \\
RoentGen         &13.1953        & 4.5612         & & 13.4895 & 5.6218\\
UniXGen          & 14.2389      & 6.2578        & & 18.1956 & 5.2819 \\
LLM-CXR          &10.8494       & 4.1574         & &9.3857  & 4.0284    \\ \midrule
Ours             & \textbf{7.2489}       & \textbf{3.496}          & &\textbf{8.5383}   &\textbf{2.9828}   \\ \bottomrule
\end{tabular}
\label{tab:sota}
\end{table}

\begin{figure*}[ht]
    \centering
    \includegraphics[width=0.9\linewidth]{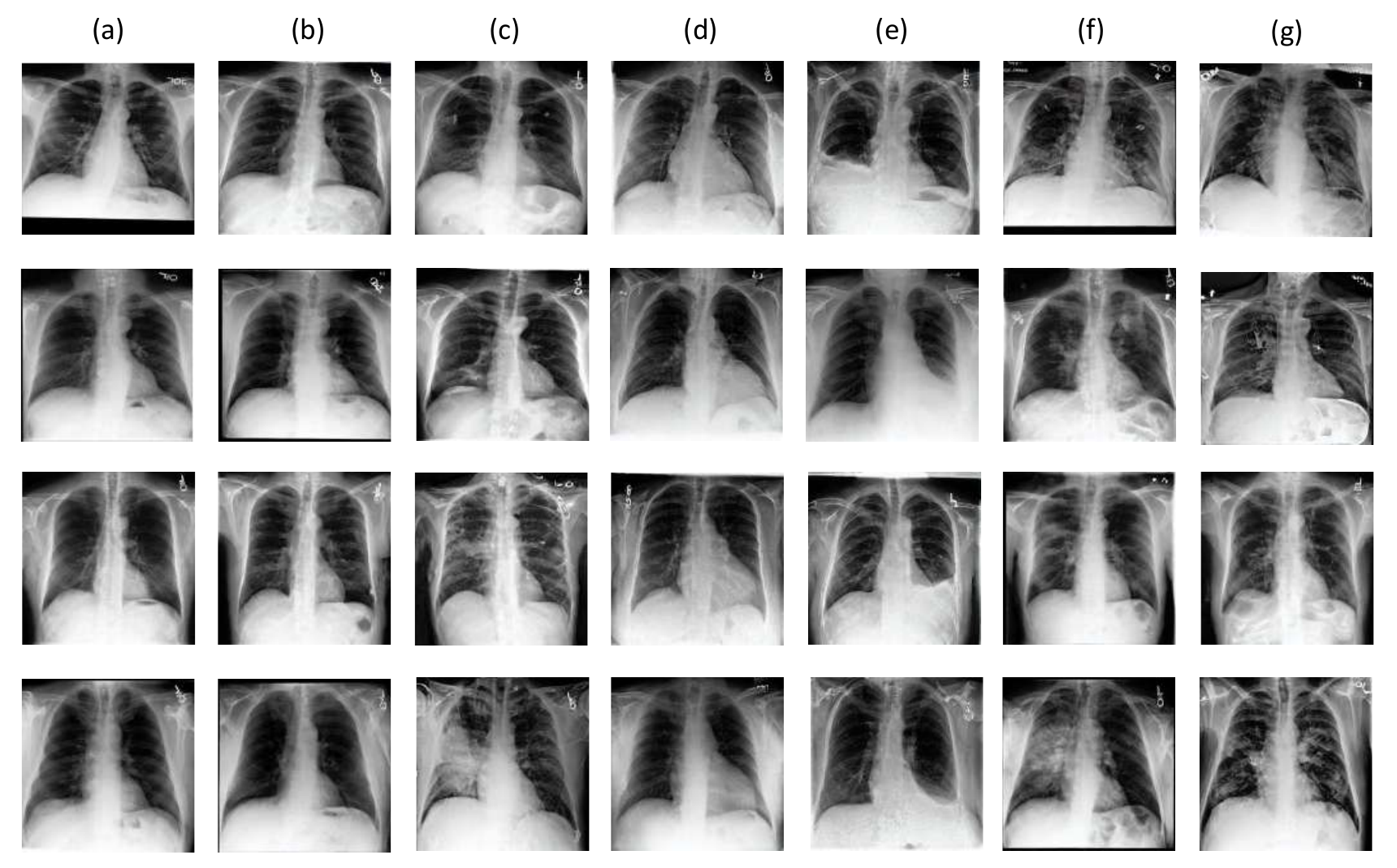}
    \caption{Counterfactual medical image generation results. Each row corresponds to CXR images from four different patients. Column (a) represents the original images, while columns (b) to (g) display the synthesized images with different pathological conditions: no finding, atelectasis, cardiomegaly, pleural effusion, consolidation and edema, respectively. The counterfactual images preserve patient identity while exhibiting distinct pathological characteristics.}
    \label{fig:VIS}
\end{figure*}

In the counterfactual generation training stage, we utilize BiomedCLIP~\cite{biomedclip} to encode medical reports into condition embeddings $e_t$, guiding the latent diffusion model to generate pathology-relevant features. We train the LDM for 100 epochs on the IU-Xray and MIMIC-CXR datasets for report-to-image generation, which takes around 34 hours under the same hardware configuration. The diffusion model operates in the latent space of size (1, 64, 64), where it generates pathological features $z^n$, which are later fused with identity features $z^{1:n-1}$ before being decoded back to 256 × 256 resolution images.

For identity consistency and pathology supervision, we utilize VGG-16 \cite{VGG} and CXR-CLIP \cite{CXRCLIP} encoders, respectively. Both models require input images resized to 224 × 224 before feature extraction. The identity loss is computed in the VGG-16 feature space, ensuring that identity-related components remain unchanged, while the pathology loss is computed using CXR-CLIP embeddings, enforcing pathology consistency.

During inference, we select ten reference images from the specific category to optimize the initial noise $z^n_T$ using $L_\text{P}$. The optimization stops if $L_\text{P}$ does not significantly decrease in the last five epochs after stabilization. This process ensures that the generated image aligns with the semantic characteristics of the target category. The entire optimization process takes approximately 2-3 minutes per sample.

\subsection{Comparison with State-of-the-Art}
We conducted a series of experiments to compare our method with state-of-the-art text-to-image generation methods. Our method is evaluated against four baselines: RoentGen~\cite{roentgen}, a Stable Diffusion-based model generating CXR images from text descriptions; UniXGen~\cite{unix}, a model specifically designed for generating CXR images and reports; LLM-CXR~\cite{Lee_2023}, a large language model-guided approach for synthesizing CXR images and corresponding pathology descriptions; and the original Stable Diffusion model fine-tuned on the MIMIC-CXR and IU-Xray datasets.

As shown in Table.~\ref{tab:sota}, our method achieves the lowest FID scores~\cite{FID} and NIQE scores~\cite{NIQE} on both the MIMIC-CXR and IU-Xray datasets. These results demonstrate the effectiveness of our approach in CXR image synthesis conditioned on text prompts. By explicitly disentangling pathological features, our approach significantly enhances the accuracy of lesion characteristics in generated images while preserving anatomical structures. Furthermore, the long-tail optimization module boosts performance on long-tail categories, ensuring better consistency and reliability in generated images.

Fig.~\ref{fig:sota} presents a qualitative comparison of our method with SOTA approaches on the MIMIC-CXR dataset. Our method successfully simulates pathologies described in input text prompts while maintaining structural integrity. Unlike existing methods, our model effectively preserves anatomical features while achieving diverse and accurate pathology synthesis, making it highly effective for counterfactual image generation.

\subsection{Semantic Analysis}
\begin{table*}[ht]
\centering
\caption{Comparison of the proposed method with state-of-the-art methods  on F1 Score and AUROC in medical image generation tasks.}
\vspace{-3mm}
\label{tab:f1_auc}
\begin{tabular}{lcccccccccccc}
\toprule
\textbf{F1-score} & \textbf{Atel.} & \textbf{Cnsl.} & \textbf{Pmtx.} & \textbf{Edema} & \textbf{Eff.} & \textbf{Pna.} & \textbf{Cmgl.} & \textbf{Les.} & \textbf{Frac.} & \textbf{Opac.} & \textbf{ECm.} & \textbf{Avg.} \\
\midrule
RoentGen & 0.7860 & 0.7067 & 0.6897 & 0.6266 & 0.7390 & 0.4681 & 0.7410 & 0.5616 & 0.6963 & 0.7248 & 0.6924 & 0.6757 \\
UniXGen  & 0.8398 & 0.6996 & 0.5231 & 0.7587 & 0.6868 & 0.6397 & 0.7340 & 0.5438 & 0.6231 & 0.7560 & 0.7519 & 0.6870 \\
LLM-CXR  & 0.8313 & 0.7835 & 0.6813 & \textbf{0.7819} & 0.7937 & 0.7304 & \textbf{0.8151} & 0.7447 & 0.6921 & \textbf{0.8392} & 0.7826 & 0.7702 \\
\midrule
Ours     & \textbf{0.8591} & \textbf{0.8103} & \textbf{0.8028} & 0.7801 & \textbf{0.8174} & \textbf{0.7825} & 0.8018 & \textbf{0.7581} & \textbf{0.7857} & 0.8271 & \textbf{0.7947} & \textbf{\textcolor{blue}{0.8019}} \\
\bottomrule
\end{tabular}

\vspace{0.5cm}  

\begin{tabular}{lcccccccccccc}
\toprule
\textbf{AUROC} & \textbf{Atel.} & \textbf{Cnsl.} & \textbf{Pmtx.} & \textbf{Edema} & \textbf{Eff.} & \textbf{Pna.} & \textbf{Cmgl.} & \textbf{Les.} & \textbf{Frac.} & \textbf{Opac.} & \textbf{ECm.} & \textbf{Avg.} \\
\midrule
RoentGen & 0.7408 & 0.7821 & 0.7256 & 0.6942 & \textbf{0.8155} & 0.5933 & 0.7324 & 0.6157 & 0.5926 & 0.7318 & 0.7848 & 0.7099 \\
UniXGen  & 0.7822 & 0.7359 & 0.6507 & 0.7718 & 0.7360 & 0.7024 & \textbf{0.8068} & 0.7056 & 0.6979 & 0.7710 & 0.8035 & 0.7422 \\
LLM-CXR  & 0.7893 & 0.8098 & 0.7389 & 0.7949 & 0.7992 & 0.7467 & 0.7689 & 0.7695 & 0.7244 & \textbf{0.8545} & 0.8168 & 0.7830 \\
\midrule
Ours     & \textbf{0.8556} & \textbf{0.8304} & \textbf{0.7868} & \textbf{0.8089} & 0.8125 & \textbf{0.8012} & 0.7748 & \textbf{0.8375} & \textbf{0.7416} & 0.8473 & \textbf{0.8196} & \textbf{\textcolor{blue}{0.8106}} \\
\bottomrule
\end{tabular}
\end{table*}

\begin{table}[ht]
\centering
\caption{Results of expert assessment conducted by ten radiologists evaluating the generated images, focusing on pathological classification accuracy and anatomical structure consistency.}
\vspace{-3mm}
\label{tab:exp}
\begin{tabular}{>{\centering\arraybackslash}m{3cm}cc}
\toprule
\textbf{\makecell[c]{Disease Category}} & \textbf{\makecell[c]{Pathological \\ Classification}} & \textbf{\makecell[c]{Anatomical \\ Structure}} \\ \midrule
Atelectasis        & 0.93 ± 0.05 & 4.27 ± 0.20 \\
Consolidation        & 0.82 ± 0.12 & 3.96 ± 0.26 \\
Edema        & 0.88 ± 0.07 & 4.39 ± 0.29 \\
Pneumonia         & 0.84 ± 0.04 & 4.41 ± 0.36 \\
Pneumothorax         & 0.81 ± 0.07 & 4.02 ± 0.42 \\ \midrule
\textbf{Overall}  &  0.86 ± 0.09 & 4.21 ± 0.37 \\ 
\bottomrule
\end{tabular}
\end{table}

To demonstrate the performance of our method in pathological simulation, we conduct multi-label pathology classification and evaluated the classification performance on both real and synthetic images. 
As shown in the Table~\ref{tab:f1_auc} , we measure the accuracy of pathologies of generated images by calculating AUROC and F1 scores against the images generated by SOTA works. We used a DenseNet-121\cite{densenet} model pretrained on MIMIC-CXR dataset for CXR pathologies classification. 
The results demonstrate that our method achieves better alignment between textual descriptions and generated images compared to the baselines, as reflected by higher AUROC and F1 scores. This indicates that our approach not only generates high-quality images but also ensures that the visual features of generated images align closely with the input textual descriptions. Furthermore, by leveraging the long-tail optimization, our method exhibits improved performance on long-tail categories such as consolidation and pneumothorax. Additionally, the experiment demonstrates that this optimization does not affect the image synthesis performance on common categories. For these categories, our method is able to achieve comparable or better results.

Fig.~\ref{fig:VIS} presents additional counterfactual medical images generated by our method. By leveraging different text prompts to guide the synthesis of pathological features, our approach enables the generation of diverse medical images while maintaining patient identity. This not only enhances the interpretability of the model but also provides valuable clinical insights, making it a useful tool for medical analysis and decision-making.
\subsection{Expert Assessment}
To further validate the effectiveness of our method, we conducted an expert evaluation experiment, generating a total of 100 CXR images across five pathological categories, with 20 images per category. Additionally, each generated image was paired with its corresponding original image, forming 100 paired image samples for comparative assessment. Ten experienced radiologists were invited to evaluate the generated images based on two key criteria: pathological classification accuracy and anatomical structure preservation.
\begin{itemize}
    \item \textbf{Pathological Classification Accuracy}: This metric evaluates whether the generated images correctly simulate pathological features. The experts assessed whether the synthesized pathological features matched the expected condition. A score of 1 was assigned if the pathology was correctly generated, and 0 otherwise.
    
    \item \textbf{Anatomical Structure Preservation}: This metric assesses whether the causal counterfactual generation approach maintains the key anatomical structures while synthesizing pathological features. A 0-5 scale was used for scoring, with higher scores indicating better structural consistency with the original images.
\end{itemize}

Table~\ref{tab:exp} presents the mean and standard deviation of the expert evaluation scores. Our method demonstrated high accuracy and stability in counterfactual pathological generation, achieving an average pathological classification accuracy of 86\% across all categories. Notably, even for long-tail categories (e.g., consolidation and pneumothorax), our model maintained comparable performance to common categories (e.g., atelectasis and pulmonary edema), verifying its robustness and generalization ability.

Furthermore, the anatomical structure preservation scores remained consistently high across all categories, with an average score of 4.21/5, indicating that our method effectively retains essential anatomical features. This result underscores the advantage of our causal disentanglement strategy, ensuring that the original structural integrity is preserved while synthesizing pathology-specific features.


\begin{table}[ht]
\centering
\renewcommand{\arraystretch}{1.2} 
\setlength{\tabcolsep}{2.5pt} 
\caption{Ablation study results evaluating the impact of optimization modules (Group Supervision, Transformer, Medical Report Refinement, Long-tail Category Optimization) on image generation quality (FID) and semantic accuracy (mAUC)  to the VAE disentangling baseline.}
\vspace{-3mm}
\label{tab:abl}
\begin{tabular}{ccccccc}
\toprule
\multicolumn{5}{c}{\textbf{Method}} & \multicolumn{2}{c}{\textbf{Metrics}} \\
\cmidrule(lr){1-5} \cmidrule(lr){6-7} 
\textbf{Baseline} & \textbf{$L_\text{ID}+L_\text{PC}$} & \textbf{$\mathcal{T}$} & \textbf{GPT} & \textbf{$L_\text{P}$} & \textbf{FID} & \textbf{mAUC} \\
\midrule
$\checkmark$ &           &           &     &            & 22.8242 & 0.5918 \\
$\checkmark$ & $\checkmark$ &           &     &            & 20.4751 & 0.6138 \\
$\checkmark$ &           & $\checkmark$ &     &            & 13.2748 & 0.7326 \\
$\checkmark$ & $\checkmark$ & $\checkmark$ &     &            & 11.9832 & 0.7684 \\
$\checkmark$ & $\checkmark$ & $\checkmark$ & $\checkmark$ &     & 8.2359 & 0.7762 \\
$\checkmark$ & $\checkmark$ & $\checkmark$ & $\checkmark$ & $\checkmark$ & \textbf{7.2489} & \textbf{0.8106} \\
\bottomrule
\end{tabular}
\end{table}
\begin{figure}[ht]
    \centering
    \includegraphics[width=0.9\linewidth]{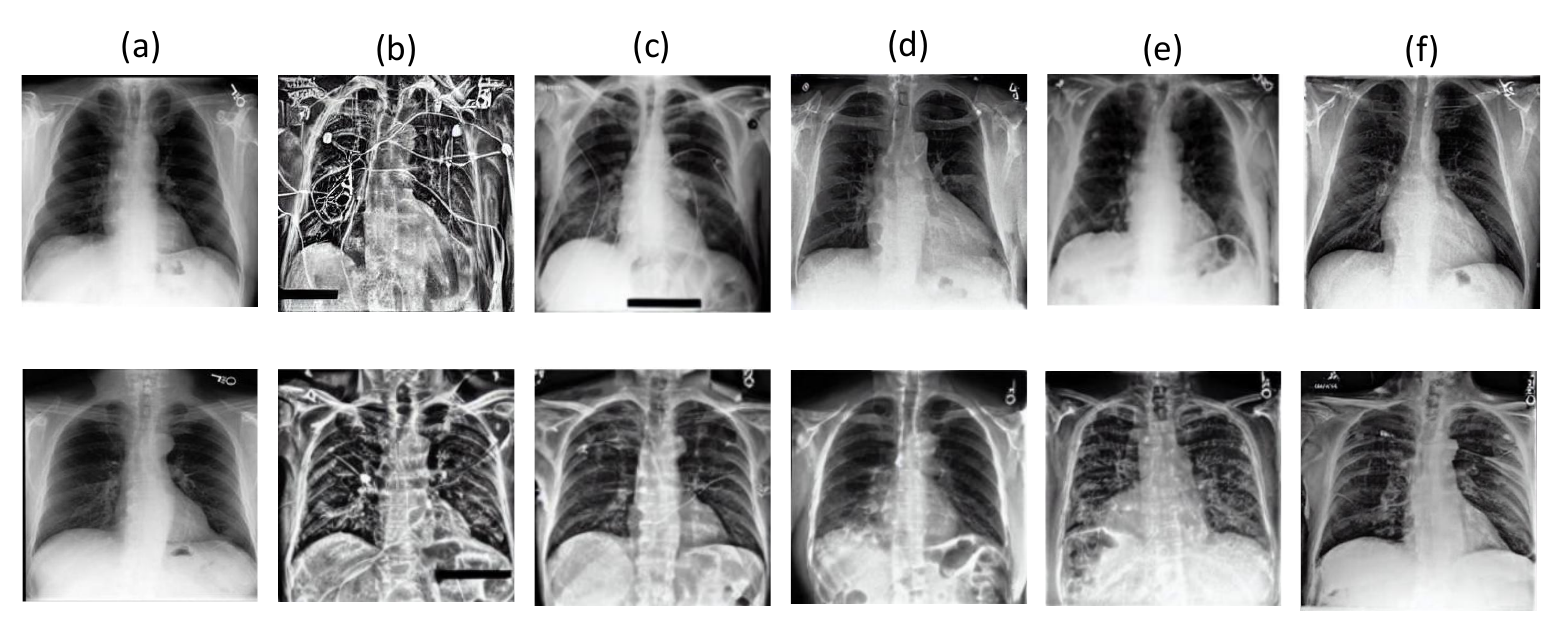}
    \caption{ Qualitative visualization of the ablation study. The figure presents a qualitative comparison of counterfactual CXR generation for two patients under cardiomegaly (top row) and consolidation (bottom row) conditions. (a) Original images, (b) VAE disentanglement baseline, (c) VAE + Transformer + group supervision, (d) Our method without group supervision, (e) Our method without long-tail optimization, (f) Our method. The results demonstrate the impact of each module on image quality and pathological feature accuracy.}
    \label{fig:able}
\end{figure}

Overall, the high agreement among expert assessments confirms the reliability of our method for clinical applications. These findings suggest that our approach not only generates realistic and diverse counterfactual medical images but also holds potential for medical training, diagnostic enhancement, and model interpretability.

\subsection{Ablation Study}
\subsubsection{Module Effectiveness Evaluation}

We conducted a comprehensive ablation study to validate the effectiveness of each module in our proposed approach. Table~\ref{tab:abl} quantitatively evaluates the quality and semantic accuracy of the generated images, where a higher mAUC value indicates that the image retains more accurate pathological information. Fig.~\ref{fig:able} presents the qualitative evaluation results, where (a) represents the original images. Experimental results demonstrate that each module contributes positively to improving image quality and the accuracy of lesion information, which we will discuss in detail below.

The causal disentanglement module enables the decoupling of structural features from pathological features, allowing lesion mechanisms to act on a designated lung structure—this serves as the foundation for counterfactual generation in our method. From the FID and mAUC results, we observe that using only a VAE for disentanglement leads to instability. Without additional supervision, we found that many generated images exhibited excessive noise across the entire image, as shown in Fig.~\ref{fig:able}(b). This is because the model fails to capture the causal relationship between pathological features and structural characteristics, which negatively affects image generation. Additionally, the presence of redundant information in medical reports further degrades the quality of the generated images. After incorporating the multi-layer Transformer module, the model learns the interactions between different influencing factors more effectively, leading to enhanced stability in image generation, as demonstrated in column (c) of Fig.~\ref{fig:able}. Moreover, training with processed medical reports reduces the proportion of unstable images.

To assess the importance of group-supervised training, we removed it from the training process and observed the results. While the generated images still maintained structural stability, their anatomical structures deviated from the original images, as seen in column (d) of Fig.~\ref{fig:able}. This is likely due to the lack of supervision, which prevents the model from ensuring that the extracted $z^{1:n-1}$ truly represents the patient’s identity information. Such instability in training ultimately degrades model performance, indicating that group supervision plays a critical role in ensuring the counterfactual nature of disentanglement.

Without long-tail optimization, the model has difficulty in generating clear or accurate pathological features for specific prompts, as illustrated in column (e) of Fig.~\ref{fig:able}. Then, Fig.~\ref{fig:able} (f) shows that aligning the semantic information between reference images of long-tail categories and generated images enables the model to synthesize these categories more reliably. Furthermore, since this optimization process does not require additional training, it does not affect the model's performance on common categories. This highlights its high applicability in medical image synthesis tasks.

\subsubsection{Discussion on Disentangled Vectors and Transformer}
\begin{table*}[t]
    \centering
    \caption{Counterfactual data augmentation leads to significant improvements in multi-label classification tasks}
    \label{tab:class}
    \begin{tabular}{@{\hskip 5pt}llccccccc@{\hskip 5pt}}
        \toprule
        \textbf{Category} & \textbf{Model} & \textbf{AUC $\uparrow$} & \textbf{Accuracy $\uparrow$} & \textbf{Sensitivity $\uparrow$} & \textbf{Specificity $\uparrow$} & \textbf{PPV $\uparrow$} & \textbf{NPV $\uparrow$} & \textbf{F1 Score $\uparrow$} \\
        \midrule
        \multirow{4}{*}{Atelectasis} 
                                & None        & 73.80 $\pm$ 1.27 & 68.39 $\pm$ 0.24 & 68.75 $\pm$ 1.24 & 69.71 $\pm$ 0.43 & \textbf{48.58 $\pm$ 0.25} & 86.72 $\pm$ 0.17 & 59.14 $\pm$ 0.44 \\
                                & SD      & 72.60 $\pm$ 0.75 & 70.21 $\pm$ 0.43 & 72.16 $\pm$ 0.29 & 71.64 $\pm$ 0.31 & 46.10 $\pm$ 0.70 & \textbf{94.88 $\pm$ 0.50} & 56.90 $\pm$ 0.85 \\
                                & Ours-w/o $L_p$  & \textbf{74.26 $\pm$ 0.65} & 72.35 $\pm$ 0.91 & 72.32 $\pm$ 0.32 & \textbf{72.18 $\pm$ 0.45} & 45.75 $\pm$ 0.86 & 89.55 $\pm$ 0.31 & \textbf{\textcolor{blue}{59.24 $\pm$ 0.62}} \\
                                & Ours    & 74.10 $\pm$ 0.60 & \textbf{73.20 $\pm$ 0.38} & \textbf{74.16 $\pm$ 0.37} & 72.16 $\pm$ 0.43 & 46.15 $\pm$ 0.76 & 89.53 $\pm$ 0.26 & 59.13 $\pm$ 0.48 \\
        \midrule
        \multirow{4}{*}{Cardiomegaly} 
                                & None       & \textbf{81.76 $\pm$ 0.68} & \textbf{82.75 $\pm$ 0.41} & 75.88 $\pm$ 1.45 & \textbf{81.08 $\pm$ 0.43} & 59.38 $\pm$ 1.03 & 95.82 $\pm$ 0.14 & 64.46 $\pm$ 1.18 \\
                                & SD      & 81.00 $\pm$ 0.40 & 76.47 $\pm$ 0.31 & 76.42 $\pm$ 1.20 & 77.56 $\pm$ 0.38 & 54.99 $\pm$ 1.03 & 90.78 $\pm$ 0.29 & 67.34 $\pm$ 1.08 \\
                                & Ours-w/o $L_p$  & 81.68 $\pm$ 0.51 & 80.35 $\pm$ 0.45 & 75.75 $\pm$ 1.20 & 81.04 $\pm$ 0.43 & 58.99 $\pm$ 1.12 & 95.83 $\pm$ 0.25 & 67.33 $\pm$ 0.92 \\
                                & Ours    & 81.72 $\pm$ 0.55 & 79.97 $\pm$ 0.37 & \textbf{76.42 $\pm$ 1.03} & 81.00 $\pm$ 0.41 & \textbf{60.10 $\pm$ 1.05} & \textbf{95.89 $\pm$ 0.21} & \textbf{\textcolor{blue}{67.54 $\pm$ 1.07}} \\
        \midrule
        \multirow{4}{*}{Consolidation} 
                                & None        & 69.26 $\pm$ 0.53 & 63.15 $\pm$ 0.43 & 74.95 $\pm$ 1.04 & 72.04 $\pm$ 0.46 & 37.83 $\pm$ 0.84 & 86.06 $\pm$ 0.17 & 36.20 $\pm$ 0.92 \\
                                & SD      & 73.12 $\pm$ 0.53 & 67.56 $\pm$ 0.41 & \textbf{75.65 $\pm$ 1.04} & 75.17 $\pm$ 0.42 & 41.57 $\pm$ 0.91 & 86.72 $\pm$ 0.27 & 41.51 $\pm$ 0.91 \\
                                & Ours-w/o $L_p$  & 71.02 $\pm$ 0.40 & 70.94 $\pm$ 0.49 & 75.03 $\pm$ 0.83 & \textbf{75.90 $\pm$ 0.41} & 43.37 $\pm$ 0.86 & \textbf{88.32 $\pm$ 0.37} & 42.11 $\pm$ 0.76 \\
                                & Ours    & \textbf{74.45 $\pm$ 0.43} & \textbf{74.85 $\pm$ 0.41} & 75.30 $\pm$ 0.42 & 75.16 $\pm$ 0.41 & \textbf{46.08 $\pm$ 0.92} & 87.08 $\pm$ 0.30 & \textbf{\textcolor{blue}{43.16 $\pm$ 0.69}} \\
        \midrule
        \multirow{4}{*}{Pneumonia} 
                                & None       & 73.60 $\pm$ 1.21 & 73.12 $\pm$ 0.43 & 68.04 $\pm$ 0.53 & 75.34 $\pm$ 0.43 & 30.63 $\pm$ 0.32 & 89.34 $\pm$ 0.11 & \textbf{\textcolor{blue}{56.19 $\pm$ 0.61}} \\
                                & SD      & 73.56 $\pm$ 2.71 & \textbf{73.71 $\pm$ 0.43} & 67.54 $\pm$ 0.43 & \textbf{77.93 $\pm$ 0.32} & 38.33 $\pm$ 0.17 & 89.20 $\pm$ 0.31 & 53.89 $\pm$ 0.51 \\
                                & Ours-w/o $L_p$  & \textbf{77.53 $\pm$ 1.74} & 72.71 $\pm$ 0.45 & 67.76 $\pm$ 0.16 & 77.76 $\pm$ 0.16 & 39.61 $\pm$ 0.24 & 89.10 $\pm$ 0.19 & 55.78 $\pm$ 0.51 \\
                                & Ours    & 77.12 $\pm$ 1.84 & 72.93 $\pm$ 0.35 & \textbf{68.07 $\pm$ 1.42} & 77.90 $\pm$ 0.68 & \textbf{40.77 $\pm$ 0.17} & \textbf{89.47 $\pm$ 0.32} & 55.48 $\pm$ 0.42 \\
        \midrule
        \multirow{4}{*}{Pneumothorax} 
                                & None     & 70.81 $\pm$ 1.41 & 64.01 $\pm$ 0.35 & 68.08 $\pm$ 4.30 & 69.40 $\pm$ 1.60 & 29.63 $\pm$ 0.40 & 90.31 $\pm$ 0.15 & 39.49 $\pm$ 1.18 \\
                                & SD      & 73.90 $\pm$ 2.11 & 67.60 $\pm$ 0.42 & 71.23 $\pm$ 1.72 & \textbf{69.70 $\pm$ 0.81} & 38.25 $\pm$ 0.38 & 91.60 $\pm$ 0.15 & 51.55 $\pm$ 0.38 \\
                                & Ours-w/o $L_p$  & 73.53 $\pm$ 1.38 & 67.14 $\pm$ 0.46 & 72.12 $\pm$ 1.38 & 67.14 $\pm$ 0.38 & 38.69 $\pm$ 0.51 & 87.16 $\pm$ 0.11 & 51.07 $\pm$ 0.91 \\
                                & Ours    & \textbf{73.98 $\pm$ 1.28} & \textbf{69.14 $\pm$ 0.42} & \textbf{72.19 $\pm$ 0.32} & 67.82 $\pm$ 0.32 & \textbf{38.90 $\pm$ 0.21} & \textbf{92.46 $\pm$ 0.10} & \textbf{\textcolor{blue}{52.36 $\pm$ 0.25}} \\
        \bottomrule
    \end{tabular}
\end{table*}

\begin{figure}
    \centering
    \includegraphics[width=0.7\linewidth]{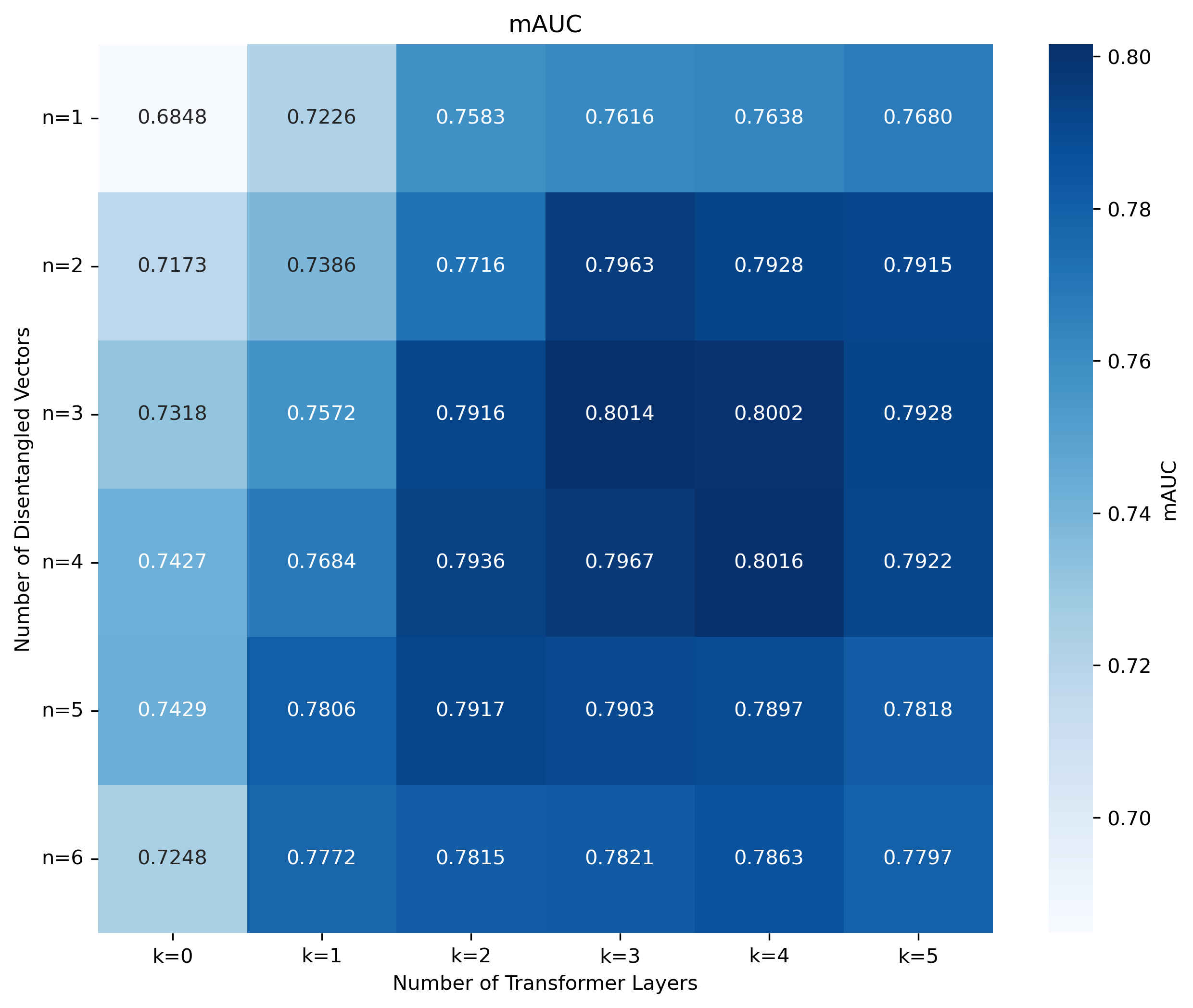}
    \caption{Confusion matrix regarding the impact of the disentangled feature vector $n$ and the number of transformer layers $k$ on the accuracy of the generated images. After adjusting $n$ and $k$, we input the generated images into the pretrained classifier to observe the accuracy of the image's pathological features.}
    \label{fig:heat}
\end{figure}
As shown in Fig.~\ref{fig:heat}, we conducted ablation experiments to analyze the impact of the number of disentangled features $n$ and the number of Transformer layers $k$ on semantic feature accuracy. To quantitatively evaluate the performance, we utilized a DenseNet-121 model pretrained on the MIMIC-CXR dataset to assess the mAUC of images generated by our method on the test set. The figure presents the experimental results, where $n$ represents the number of disentangled feature channels, and $k$ denotes the number of Transformer layers. Specifically, when $n=1$, no feature disentanglement is performed, and 
the latent diffusion model learns the features of the entire image without structural separation.

Through extensive comparative experiments, we found that the best performance is achieved when $n=4$ and $k=4$, balancing both feature decoupling effectiveness and model stability.

Our counterfactual generation approach relies on the model’s ability to effectively disentangle pathological features from structural features. To examine the influence of disentangling, we set $n=2$ as the baseline for analysis. When $n=2$, the model has difficulty to accurately separate the two highly coupled feature types due to their strong interdependence. Increasing the number of disentangled vectors to $n=4$ mitigates this issue by providing greater flexibility in feature separation, allowing the model to better isolate pathology from identity-related characteristics.

However, further increasing $n$ does not consistently improve results. Since our loss function primarily constrains the validity of pathological features and the independence of disentangled vectors without explicit constraints on other components, an excessive number of disentangled vectors can lead to instability during training. The lack of direct supervision for additional feature channels makes their representations less controllable, leading to increased training difficulty. Given the limited annotations available in medical imaging, adding more disentangled vectors beyond $n=4$ does not yield better generation quality and may introduce unnecessary noise.

Similarly, an appropriate number of Transformer layers enhances the model’s ability to capture inter-feature relationships. When $k=4$, the model effectively learns the causal interactions between features, leading to optimal performance. However, when $k$ is increased to 9, we observe a decline in performance. This is likely due to vanishing gradients, which make training deeper Transformers more challenging and ultimately hinder convergence, reducing the generalizability of the generated images.

These findings confirm that setting $n=4$ and $k=4$ provides the best trade-off between feature disentanglement, training stability, and semantic accuracy in counterfactual medical image generation.

\subsection{Downstream Task}
We designed a downstream task experiment to validate the effectiveness of our medical image generation method in data augmentation for multi-class models.
As shown in Table.~\ref{tab:class}, we trained a ResNet18 \cite{resnet} model on the MIMIC-CXR dataset and evaluated its performance across five representative pathological categories. The “None” row in the results table represents the model trained only on the original training dataset. To assess the effectiveness of our generation method, we augmented the dataset with synthetic data generated by three approaches: the original diffusion model (SD), our proposed method (Ours), and our method without the long-tail optimization module (Ours-w/o $L_\text{P}$). The dataset was proportionally augmented based on class labels and used to retrain ResNet18. Results show that synthetic data effectively improve training performance for long-tail categories. Our method outperforms conventional diffusion models by enabling a broader data distribution, and the long-tail optimization module $L_\text{P}$ enhances the stability of synthetic images for long-tail categories without affecting the quality of synthesis for common categories. However, for common categories such as cardiomegaly and atelectasis, the already high performance is slightly impacted by the noise introduced during data augmentation.

\section{Conclusion}
In this paper, we propose an effective medical image generation framework to address the issue of imbalanced data distribution. By introducing a causal disentanglement mechanism, our method models the interactions between identity-related and pathology-related features, enabling text-guided generation of diverse and structurally consistent counterfactual images. This process effectively augments the data distribution by preserving identity while synthesizing pathology. Furthermore, to tackle the challenge of generation for long-tail categories, we incorporate a training-free optimization strategy based on latent diffusion models, improving generation quality in rare categories. Experimental results demonstrate that our approach accurately captures the interplay between identity and pathology and achieves superior performance in both counterfactual image synthesis and downstream data augmentation.

\section{Acknowledgments}
Sincere gratitude to KeLiang Xie from Tianjin Medical University General Hospital and He Jiao from Tianjin Chest Hospital for their  guidance in the medical field and assistance in organizing expert assessment.
This work was supported in part by the National Natural Science
Foundation of China (62272337) and the Natural Science Foundation of Tianjin (16JCZDJC31100).

\normalem
\bibliographystyle{IEEEtran}
\bibliography{ref}

\ifCLASSOPTIONcaptionsoff
  \newpage
\fi





%
\end{document}